\documentclass[letterpaper]{article} 
\usepackage[preprint]{aaai2027}  
\usepackage[hyphens]{url}  
\usepackage{graphicx} 
\usepackage{natbib}  
\usepackage{caption} 
\usepackage{algorithm}
\usepackage{algorithmic}
\usepackage{amsmath}
\usepackage{amssymb}
\usepackage{siunitx}
\usepackage{booktabs}
\usepackage{comment}
\usepackage{cuted}

\usepackage{newfloat}
\usepackage{listings}
\DeclareCaptionStyle{ruled}{labelfont=normalfont,labelsep=colon,strut=off} 
\floatstyle{ruled}
\newfloat{listing}{tb}{lst}{}
\floatname{listing}{Listing}
\usepackage{tcolorbox}
\tcbuselibrary{skins, breakable, theorems}
\usepackage{colortbl}
\usepackage{makecell}
\usepackage{hhline}
\usepackage{array}
\usepackage{subcaption}
\usepackage{multirow}

\usepackage{relsize}
\title{ICI-VLA: In-Context Imitation with Spatiotemporally Aligned Demonstrations for Vision-Language-Action Models}

\author{
    Songhua Yang\textsuperscript{\rm 1},
    Ziyu Liu\textsuperscript{\rm 1},
    Xuetao Li\textsuperscript{\rm 1},
    Ruqi Xiao\textsuperscript{\rm 1},
    Kangxin Zhu\textsuperscript{\rm 1},
    Miao Li\textsuperscript{\rm 2}
}
\affiliations{
    \textsuperscript{\rm 1}Wuhan University, Wuhan, China\\
    \textsuperscript{\rm 2}Institute of Technological Sciences, Wuhan University, Wuhan, China
}

\begin{document}

\maketitle

\begin{abstract}
Vision-Language-Action (VLA) policies are commonly adapted to new manipulation settings through additional gradient updates, which limits rapid deployment when task-specific data or compute is scarce. We present ICI-VLA, a training and retrieval framework that equips a text-action VLM with few-shot test-time adaptation through in-context demonstrations. Unlike mainstream VLA designs based on action-specific multimodal fusion, ICI-VLA retains the native text-generation interface. ICI-VLA updates its parameters only during offline training; at inference, the policy remains fixed and conditions action generation on retrieved micro-demonstrations. The framework decomposes long trajectories into short, semantically labeled examples and trains an RD-Encoder with positives mined by Dynamic Time Warping (DTW), aligning the retrieved context with the phase and geometry of the current subtask. We further introduce Target Action Masking, a context-corruption objective designed to reduce direct action copying and increase reliance on the current observation. ICI-VLA reaches average success rates of 97.7\% on LIBERO and 60.4\% on RoboTwin 2.0, exceeding the highest reported baseline average on RoboTwin 2.0 by 19.3 percentage points. It also achieves 83.2\% across four physical tasks. These results indicate that a fixed VLA policy can benefit from conditioning on spatiotemporally aligned demonstrations at test time.
\end{abstract}


\section{Introduction}

Vision-Language-Action (VLA) models transfer representations learned from large-scale vision-language data to low-level robotic control \cite{brohan2023rt1, zitkovich2023rt2, kim2025openvla, physical2025pi05}. Despite strong multi-task performance, most deployed VLAs behave as fixed conditional policies: adapting their behavior typically requires collecting additional demonstrations and updating model parameters \cite{ma2024survey, gao2023kvil, goyal2025vla0}. Repeating this process for each new task is costly and limits rapid adaptation in changing environments.

In-Context Learning (ICL) offers a promising alternative to task-by-task retraining \cite{brown2020language}. By conditioning on a few demonstrations in the input, language and vision-language models can adapt their outputs to a new task without updating parameters \cite{ye2023incontext, bar2022visual, wang2023images}. Because popular VLAs build on these pretrained multimodal models, demonstrations could likewise provide task information at inference rather than serve only as data for additional fine-tuning. Transferring this capability to physical control, however, remains difficult: robotic demonstrations couple visual observations, language, proprioception, and continuous actions, and poorly matched context can interfere with action prediction rather than improve it.

\begin{figure}[!t]
  \includegraphics[width=\columnwidth]{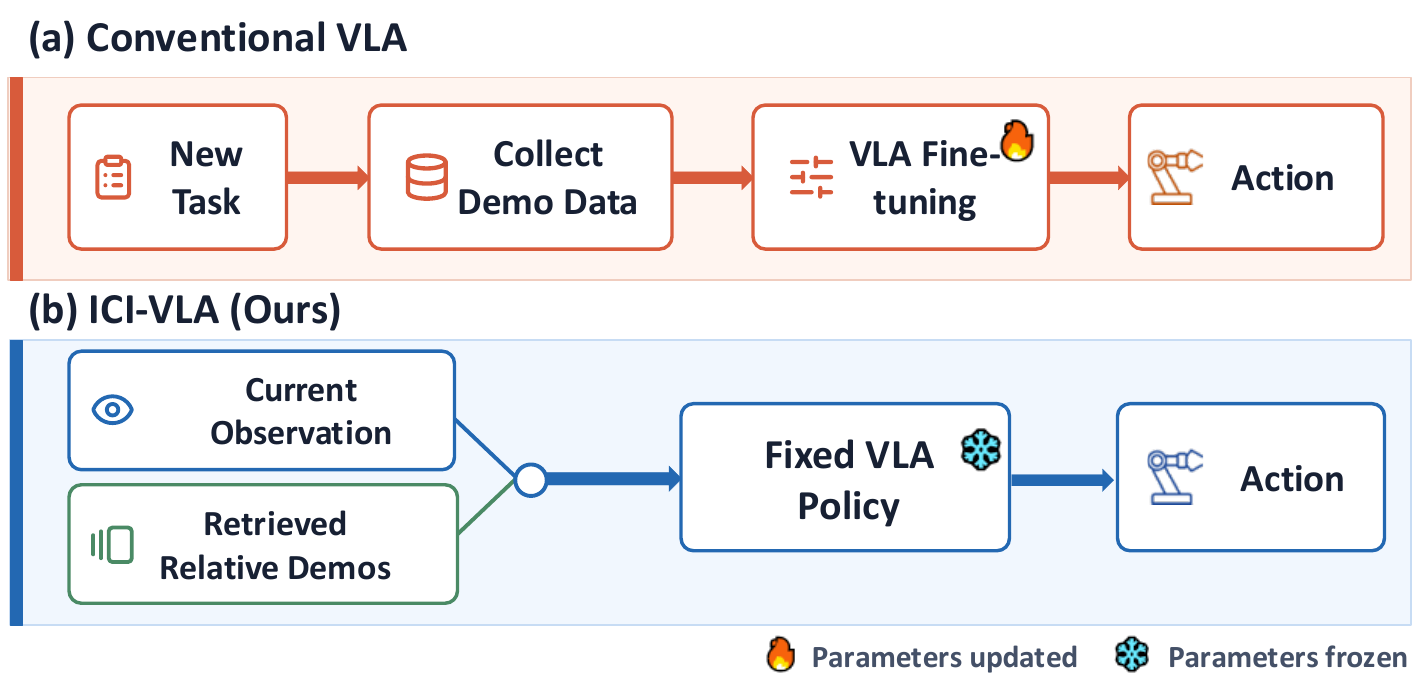}
  \caption{\textbf{Task Adaptation Paradigms.} (a) Conventional VLAs collect task-specific data and update policy parameters for each new task. (b) ICI-VLA conditions a fixed policy on the current observation and retrieved relative demonstrations, requiring no test-time gradient updates.}
  \label{fig:paradigm}
\end{figure}

Recent work has introduced retrieval and memory into generalist agents \cite{sridhar2025regent, torne2026mem}. Structured interfaces translate trajectories into language- or keypoint-based representations \cite{kwon2024language, di_palo_keypoint_2024}, while KARAG and Instant-Fold use task-aligned retrieval or single-demonstration conditioning \cite{lin2025karag, wang2026instantfold}. These results establish the feasibility of robotic ICL, but their specialized representations and task settings leave robust, general-purpose adaptation unresolved. In particular, long trajectories can be misaligned with short action chunks (\textit{temporal aliasing}), while visually or semantically similar demonstrations can require incompatible motions (\textit{semantic-dynamic mismatch}). Under either mismatch, context can become a misleading action prefix that encourages sequence copying \cite{yuan-etal-2024-llms}. Meanwhile, VLA-0 shows that a pure VLM can generate continuous actions directly as text and still achieve strong manipulation performance without an action-specific prediction head \cite{goyal2025vla0}, making its native generation interface a promising foundation for robotic ICL. 

To bridge these gaps, we introduce \textbf{ICI-VLA}, a unified framework for fine-grained in-context imitation. As illustrated in Figure \ref{fig:paradigm}, ICI-VLA replaces task-specific parameter updates with retrieval-conditioned inference: at test time, the policy remains fixed and conditions action generation on the current observation and retrieved demonstrations. To make long robotic trajectories suitable as context, the framework first decomposes full episodes into semantically labeled \textit{micro-demonstrations}, reducing the temporal mismatch between multi-stage trajectories and short action chunks. It then trains an RD-Encoder using semantic hard-filtering and Dynamic Time Warping (DTW) geometric ranking, so that retrieved contexts match both the task semantics and required motion. Finally, Target Action Masking corrupts contextual action targets during offline training, discouraging direct trajectory copying and encouraging predictions grounded in the current observation. Together, these components turn demonstrations into aligned test-time context rather than additional data for task-specific fine-tuning.

We evaluate ICI-VLA on LIBERO, RoboTwin 2.0, and four physical dual-arm Aloha tasks. ICI-VLA obtains average success rates of 97.7\% on LIBERO and 60.4\% on RoboTwin 2.0; the latter is 19.3 percentage points above the strongest reported baseline average. The physical evaluation yields an average success rate of 83.2\%. 
Together with the component ablations, these results support the effectiveness of demonstration alignment and target-action corruption under the reported benchmark protocols.

In summary, our contributions are:
\begin{itemize}
  \item We formulate VLA adaptation as few-shot test-time conditioning with a fixed text-action policy and explicitly distinguish offline training from test-time inference.
  \item We construct subtask-level micro-demonstrations and use DTW-mined supervision to train an RD-Encoder for semantic and phase-aware retrieval.
  \item We introduce Target Action Masking as a context-corruption objective and evaluate its contribution together with the retrieval components in simulation and physical deployments.
\end{itemize}

\section{Related Work}

\subsection{Vision-Language-Action Models}

VLA architectures typically use one of three action interfaces~\cite{sapkota2025vla}. Generative-action models attach continuous decoders, including diffusion- or flow-based heads, to a VLM backbone \cite{li2024roboflamingo, black2025pi0, physical2025pi05, shukor2025smolvla, bjorck2025gr00t,chen2026dfmvla}. Discrete-token models quantize continuous actions and cast control as autoregressive prediction \cite{brohan2023rt1, zitkovich2023rt2, kim2025openvla, pertsch2025fast, zhong2025vlasurvey}. Customized architectures instead introduce specialized action tokenizers, hierarchical representations, or spatial modules \cite{kim2025openvlaoft, li2025hamster, lee2025molmoact, liu2024robomamba, singh2025ogvla, qu2025spatialvla, kawaharazuka2025vlareview}.


However, these paradigms typically rely on auxiliary modules or structural alterations that elevate engineering overhead and may induce shortcut learning, potentially degrading the model's native capabilities \cite{yuan-etal-2024-llms}. Unlike earlier efforts requiring intermediate representations \cite{niu2025llarva}, VLA-0 \cite{goyal2025vla0} introduces a ``zero-modification'' paradigm by fully leveraging the native text generation interface. Given its capacity to effectively preserve the reasoning and generalization priors of pretrained models in a unified manner, we therefore adopt this minimalist paradigm in our study. Although still underexplored relative to head- and token-based designs, this unified text-generation interface offers a promising foundation for in-context action adaptation.

\subsection{In-Context Learning}



In-context learning allows a pretrained model to adapt its predictions by conditioning on demonstrations supplied in the prompt, without gradient updates for the current query~\cite{brown2020language, touvron2023llama, mao2025understanding}. Instruction-level demonstrations extend this mechanism across tasks~\cite{ye2023incontext, schoch2025good}, and visual prompting applies a related principle to image-based prediction~\cite{bar2022visual, wang2023images}. Robotic ICL is less established because its context must jointly represent perception, robot state, language, and continuous control.

Early robotic ICL methods cast control as sequence prediction or generate trajectories directly with language models~\cite{kwon2024language, niu2025llarva}. Keypoint constraints and keypoint action tokens provide more structured representations for few-shot imitation~\cite{huang2025rekep, di_palo_keypoint_2024, bar2022visual,gao2023kvil,xiong2021learning,zhang2025atk}. Retrieval- and memory-based agents instead condition decisions on demonstration libraries or interaction histories~\cite{sridhar2025regent,sridhar2025ricl,torne2026mem}. 
However, prior retrieval-conditioned VLAs do not jointly study subtask-level trajectory decomposition, motion-aware retriever supervision, and target-prefix corruption for native text-action policies. ICI-VLA combines these elements to support phase-aware retrieval and short-horizon action prediction.

\section{Methodology}


This section separates the offline learning procedure from test-time execution. Offline, we construct the demonstration library, train the retriever, and fine-tune the VLA policy. At test time, all parameters remain fixed; the system adapts its action prediction only by retrieving and conditioning on a small set of demonstrations.

\subsection{Preliminaries}



VLA models aim to map multimodal visual and linguistic inputs directly to low-level control actions. Existing VLA architectures primarily fall into three categories: 1) Generative Action Head models (e.g., SmolVLA~\cite{shukor2025smolvla}), which employ additional decoding heads like diffusion models, thereby introducing extra fine-tuning overhead and risking the degradation of the VLM's language grounding capabilities; 2) Discrete Token models (e.g., RT-2~\cite{zitkovich2023rt2}, OpenVLA~\cite{kim2025openvla}), which map continuous actions to new vocabulary tokens. This approach not only restricts action resolution due to vocabulary size limits but also risks disrupting the pre-trained semantic space; and 3) Custom Architectures (e.g., OpenVLA-OFT~\cite{kim2025openvlaoft}), which require specialized action tokenizers, complicating the training pipeline and increasing deployment costs.

In contrast, inspired by the recent advancements of VLA-0~\cite{goyal2025vla0}, we adopt a ``zero-structural-modification`` pure VLM architecture. We strictly preserve the foundational network without adding any action-specific parameters or heads, representing continuous actions entirely as numerical text strings. This minimalist paradigm mitigates the interference from external modules and maximally retains the foundation model's innate reasoning and generalization abilities, thereby providing an optimal foundation for unlocking true ICL capabilities.

\begin{figure*}[!t]
  \includegraphics[width=1.0\textwidth]{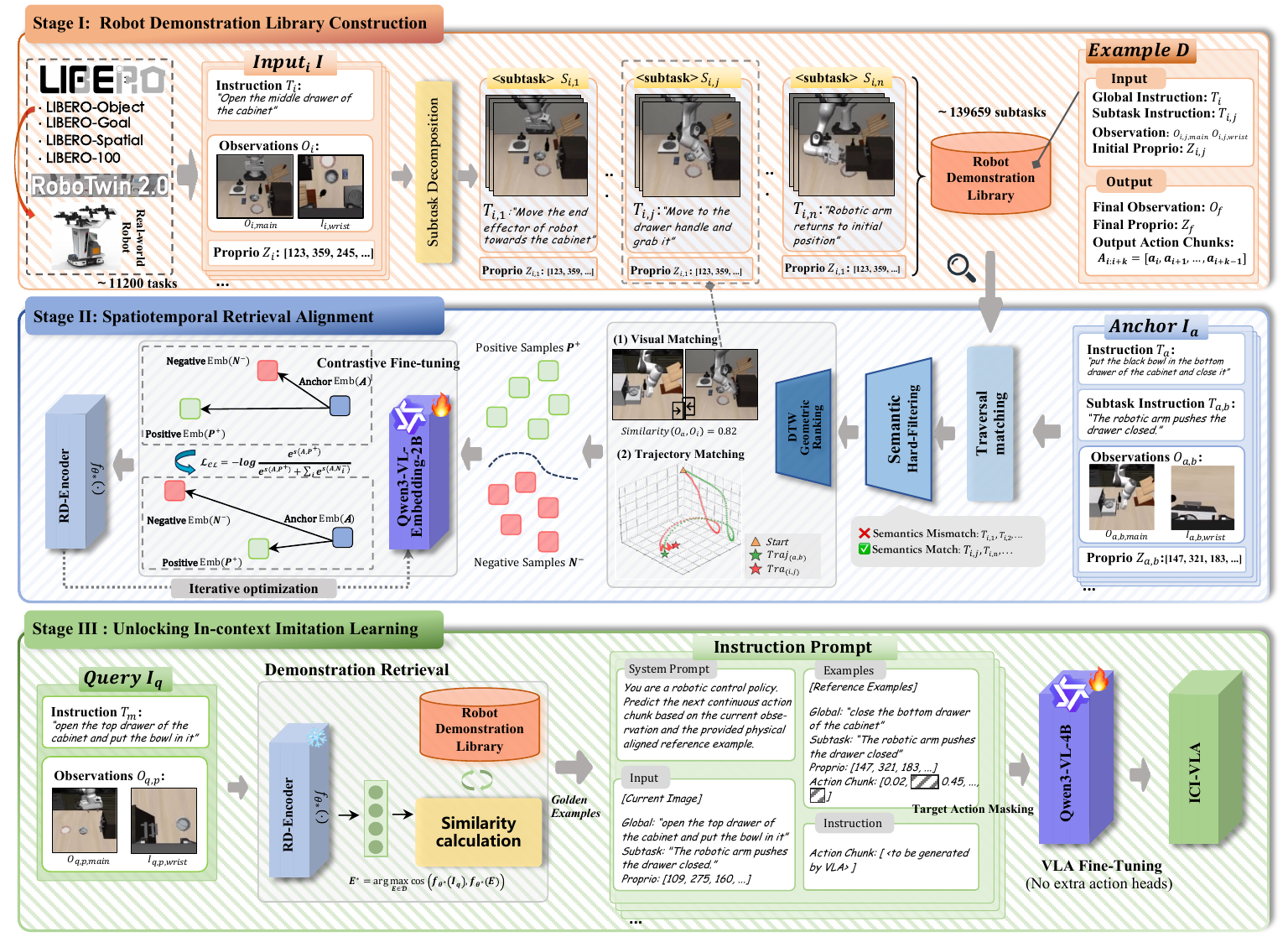}
  \caption{Overview of ICI-VLA. \textbf{(I) Library construction:} long-horizon trajectories are decomposed into fixed-horizon micro-demonstrations. \textbf{(II) Retrieval alignment:} semantic filtering and DTW-based ranking mine contrastive triplets; at inference, the frozen RD-Encoder retrieves from observable inputs without target actions. \textbf{(III) Policy training and inference:} Qwen3-VL-4B is fine-tuned without an action-specific head, while Target Action Masking reduces reliance on exact numerical continuation. Both the policy and retriever remain fixed at inference, and adaptation occurs through retrieved context.}
  \label{fig:architecture}
  \vspace{2mm}
\end{figure*}

\subsection{Problem Formulation}

ICI-VLA formulates deployment-time adaptation as conditional action generation from a phase-relevant micro-demonstration while keeping all learned parameters fixed. As illustrated in Figure~\ref{fig:architecture}, the visual planner decomposes the global instruction $T$ into an ordered sequence of subtask instructions and selects the active $T_{sub}$ from the observed execution progress, advancing when the current subtask is visually completed. At time $t$, the query is $I_q=\langle T,T_{sub},O_t,Z_t\rangle$, where $O_t=\{O_{t,main},O_{t,wrist}\}$ contains the camera observations and $Z_t$ is the proprioceptive state. Each library item is a micro-demonstration $E=\langle T_e,T_{e,sub},O_e,Z_e,A_{e:e+k}\rangle$ drawn from $\mathcal{D}$. Offline training learns the retriever parameters $\phi$ and policy parameters $\theta$; at inference,
\[
E^*=\arg\max_{E\in\mathcal{D}} s_{\phi}(I_q,E),
\qquad
A_{t:t+k}\sim\pi_{\theta}(\cdot\mid I_q,E^*),
\]
where $s_{\phi}$ is the similarity score produced by the RD-Encoder. Neither model receives gradient updates at inference, so we refer to this setting as \emph{few-shot test-time in-context adaptation}, rather than zero-shot learning.
The policy-training, retriever-training, library-construction, validation, and evaluation trajectories are non-overlapping. We follow the official task definitions of each benchmark, so task semantics, objects, layouts, or instruction templates may recur across trajectory partitions; however, no evaluation trajectory or target action is used for training or library construction.

\subsection{The ICI-VLA Framework}

Figure~\ref{fig:architecture} summarizes three stages: offline demonstration-library construction, spatiotemporal retriever training, and VLA fine-tuning with Target Action Masking. The following subsections describe each stage.


\subsubsection{\textbf{Robot Demonstration Library Construction}}


We construct $\mathcal{D}$ from approximately 11,200 long-horizon trajectories collected in LIBERO~\cite{liu2023libero}, RoboTwin 2.0~\cite{zhao2024robotwin2}, and physical deployments, including approximately 1,000 dual-arm Aloha demonstrations. To match the retrieval unit to the short action horizon, we segment each long trajectory into subtasks rather than retrieve complete episodes. As shown in Stage I of Figure~\ref{fig:architecture}, a Qwen3-VL model \cite{bai2025qwen3} heuristically segments and labels each trajectory $\tau_i$ as $\tau_i\rightarrow\{S_{i,1},S_{i,2},\dots,S_{i,n}\}$. Each subtask $S_{i,k}$ is paired with a fine-grained instruction $T_{i,k}$.

Following VLA-0~\cite{goyal2025vla0}, we represent actions and proprioceptive states as text. A micro-demonstration is
\[
E=\langle T_i,T_{i,k},O_i,Z_i,A_{i:i+k}\rangle,
\]
where $T_i$ and $T_{i,k}$ are the global and subtask instructions, $O_i$ contains the main- and wrist-camera observations, $Z_i$ is the initial proprioceptive state, and $A_{i:i+k}=[a_i,a_{i+1},\dots,a_{i+k-1}]$ is the textualized $k$-step action chunk. This process produces a precomputed library of approximately 139,659 subtask examples.



\subsubsection{\textbf{Spatiotemporal Retrieval Alignment}}



Stage II trains Qwen3-VL-Embedding-2B~\cite{li2026qwen3} as the RD-Encoder. Generic multimodal similarity can rank visually related examples that require different motions. We therefore fine-tune the encoder through iterative contrastive mining that combines semantic filtering with trajectory-based supervision.

At iteration $m$, \textbf{Semantic Hard-Filtering} uses $f_{\phi}^{(m)}$ to compute cosine similarity between an anchor $I_a$ and the library items, then retains a semantically compatible candidate pool. For example, given the subtask instruction \textit{``The robotic arm pushes the drawer closed,''} the filter removes candidates about unrelated actions and retains candidates involving drawer manipulation. DTW then ranks this reduced pool using the labeled trajectories, as described next.

After semantic filtering, DTW ranks candidate pairs using their labeled $k$-step trajectories. For anchor trajectory $\mathrm{Traj}_a$ and candidate trajectory $\mathrm{Traj}_e$, the offline mining cost is
\[
\operatorname{DTW}(\mathrm{Traj}_a,\mathrm{Traj}_e)
=\min_W\sum_{(u,v)\in W}\delta(p_{a,u},p_{e,v}),
\]
where $W$ is a valid warping path and $\delta$ measures waypoint discrepancy. The lowest-cost candidate is used as the positive $P^+$; phase-misaligned candidates from the same semantic subtask provide hard negatives $N^-$. Importantly, target trajectories and DTW are used only to construct supervision during offline retriever training. At evaluation time, the target action is unknown: the frozen RD-Encoder ranks library items from $I_q$ alone and does not compute DTW against a query action.

The RD-Encoder is trained with
\[
\mathcal{L}_{CL}=-\log
\frac{\exp(e_A^\top e_{P^+}/\tau)}
{\exp(e_A^\top e_{P^+}/\tau)+\sum_i\exp(e_A^\top e_{N_i^-}/\tau)}.
\]
Because the current embedding determines which candidates enter the mining pool, we alternate candidate retrieval, DTW-based triplet mining, and contrastive optimization. Training stops after the selected positives stabilize or the maximum number of cycles is reached. This procedure transfers the offline trajectory-based ranking signal into an encoder that can operate on observable inputs at inference.

Each waypoint contains the 3D end-effector positions of all active arms in the robot base frame, normalized by the training-split workspace bounds; $\delta$ is the mean Euclidean distance across arms. We retain the top 64 semantic candidates, use the DTW-minimal candidate as $P^+$, sample 15 phase-misaligned candidates from the same semantic subtask as negatives, and stop when fewer than $\epsilon=5\%$ of anchors change their positive or after five cycles.

\begin{table*}[ht!]
\resizebox{\textwidth}{!}{%
\begin{tabular}{l ccccc c ccc}
\toprule
\multirow{2}{*}{\textbf{Model}} & \multicolumn{5}{c}{\textbf{LIBERO}} & & \multicolumn{3}{c}{\textbf{RoboTwin 2.0}} \\
\cmidrule{2-6} \cmidrule{8-10}
 & Spatial & Object & Goal & Long & Avg. & & Easy & Hard & Avg. \\
\midrule
Octo \protect\cite{octo2024octo} & 77.5 & 87.2 & 83.1 & 49.6 & 74.4 & & 20.5 & 3.1 & 12.5 \\
OpenVLA \protect\cite{kim2025openvla} & 85.8 & 86.9 & 80.5 & 55.4 & 77.2 & & 22.0 & 3.5 & 13.5 \\
$\pi_0$-FAST \protect\cite{pertsch2025fast} & 88.6 & 87.8 & 93.5 & 74.5 & 86.1 & & 34.2 & 7.8 & 22.1 \\
MolmoAct \protect\cite{lee2025molmoact} & 88.5 & 94.1 & 88.9 & 75.8 & 86.8 & & 35.5 & 8.5 & 23.1 \\
$\pi_0$ \protect\cite{black2025pi0} & 95.5 & 97.4 & 96.9 & 83.8 & 93.4 & & 46.4 & 16.3 & 32.5 \\
$\pi_{0.5}$ - KI \protect\cite{physical2025pi05} & 96.4 & 99.1 & 96.8 & 84.5 & 94.2 & & 47.1 & 17.0 & 33.3 \\
OpenVLA-OFT \protect\cite{kim2025openvlaoft} & 96.2 & \textbf{99.5} & 96.5 & \underline{93.2} & \underline{96.4} & & \underline{55.2} & \underline{24.5} & \underline{41.1} \\
\midrule
VLA-0 \protect\cite{goyal2025vla0} & \underline{98.2} & 96.3 & \underline{97.5} & 86.1 & 94.5 & & 48.5 & 18.2 & 34.6 \\
VLA-0 w/ Naive ICL & 72.8 & 76.5 & 73.1 & 63.5 & 71.5 & & 18.2 & 2.0 & 10.7 \\
\rowcolor{gray!10} \textbf{ICI-VLA (ours)} & 
\textbf{98.5} \protect\textcolor{green!60!black}{\scriptsize{(+0.3)}} & 
\underline{98.7} {\scriptsize(-0.8)} & 
\textbf{98.0} \protect\textcolor{green!60!black}{\scriptsize{(+0.5)}} & 
\textbf{96.8} \protect\textcolor{green!60!black}{\scriptsize{(+3.6)}} & 
\textbf{97.7} \protect\textcolor{green!60!black}{\scriptsize{(+1.3)}} & & 
\textbf{72.4} \protect\textcolor{green!60!black}{\scriptsize{(+17.2)}} & 
\textbf{46.3} \protect\textcolor{green!60!black}{\scriptsize{(+21.8)}} & 
\textbf{60.4} \protect\textcolor{green!60!black}{\scriptsize{(+19.3)}} \\
\bottomrule
\end{tabular}
}
\caption{\textbf{Simulation results.} Success rate (\%); each ICI-VLA task is evaluated over 10 initialized rollouts, and suite-level entries are averaged across their constituent tasks. Best and second-best entries are \textbf{bolded} and \underline{underlined}. ``VLA-0 w/ Naive ICL'' uses the same retrieved context as ICI-VLA but omits Target Action Masking \protect\cite{goyal2025vla0}. 
Because entries from prior work follow their published benchmark protocols, the table provides a benchmark-level comparison; cross-paper differences should not be interpreted as controlled paired comparisons.
}
\label{tab:main_results}
\end{table*}

\subsubsection{\textbf{VLA Fine-Tuning via Target Action Masking}}

Stage III fine-tunes Qwen3-VL-4B to predict textualized action chunks from a structured prompt containing the system instruction, retrieved demonstrations, and current query. The RD-Encoder is frozen at this stage. The number of demonstrations can vary, although all reported ICI-VLA results use three examples unless stated otherwise.

Target Action Masking is a training-time context-corruption objective. Let $\mathcal{M}$ denote the sampled target-token positions, let $\mathcal{U}$ denote their complement, and let $\widetilde A_q=c(A_q;\mathcal{M})$ be the sequence obtained by replacing positions in $\mathcal{M}$ with \texttt{[MASK]}. We optimize only the unmasked targets:
\[
\mathcal{L}_{act}=-\sum_{j\in\mathcal{U}}
\log\pi_{\theta}\!\left(a_{q,j}\mid I_q,E^*,\widetilde a_{q,<j}\right).
\]
Thus, an unmasked token may be predicted from a prefix containing corrupted earlier action tokens. The objective directly reduces dependence on exact target-prefix continuation; it does not explicitly supervise a kinematic residual with respect to $A_e$. We hypothesize that this corruption encourages greater use of the current observation and retrieved context, and evaluate its empirical contribution through ablation. 

At inference, masking is disabled and the fine-tuned policy generates actions autoregressively with fixed parameters. Accordingly, the method performs few-shot contextual conditioning without test-time gradient updates, rather than zero-shot imitation.










\section{Experimental Evaluation}

We evaluate ICI-VLA in LIBERO, RoboTwin 2.0, and a physical dual-arm Aloha setup. The experiments measure task success, component ablations, sensitivity to context size and retriever-mining cycles, and physical-system performance. Because ICI-VLA conditions on three retrieved demonstrations at inference, we refer to this protocol as few-shot in-context adaptation.

\subsection{Experimental Setup}



We evaluate ICI-VLA in simulation and on a physical robot. LIBERO~\cite{liu2023libero} evaluates object- and spatially conditioned manipulation, while RoboTwin 2.0~\cite{zhao2024robotwin2} provides high-fidelity, long-horizon dual-arm tasks requiring precise spatiotemporal coordination. For physical evaluation, we use a dual-arm Aloha system~\cite{fu2024mobile} and approximately 1,000 teleoperated trajectories covering grasping, placing, drawer manipulation, and sorting. Together, the source data comprise approximately 11,200 long-horizon trajectories. 
Within each benchmark, trajectory partitions follow the protocol defined above, while task definitions and evaluation procedures follow the official benchmark setup.


Success rate (SR) is the primary metric. Each simulated task is evaluated over 10 rollouts with varied initial states and random seeds. All experiments are conducted on a server running Ubuntu 22.04 and equipped with eight NVIDIA A100 GPUs (80~GB). Offline, both the Qwen3-VL-Embedding-2B retriever and the Qwen3-VL-4B policy undergo full-parameter fine-tuning. The retriever uses InfoNCE with temperature $\tau=0.07$ and a global batch size of 128. The policy uses AdamW with weight decay $0.01$, a peak learning rate of $2\times10^{-5}$, linear warmup, and cosine decay. We set the action-chunk horizon to $k=25$, retrieve three examples, and run five retriever-mining cycles. During evaluation, all parameters remain fixed. The demonstration context is refreshed when the policy confidence falls below $0.65$.
\label{sec:data-protocol-placeholder}The three-shot context, five retriever-mining cycles, and confidence threshold of $0.65$ are selected exclusively on the validation split. We use two-sided 95\% Wilson score confidence intervals for binomial success rates.



\begin{figure*}[!t]
\centering
\includegraphics[width=0.99\textwidth]{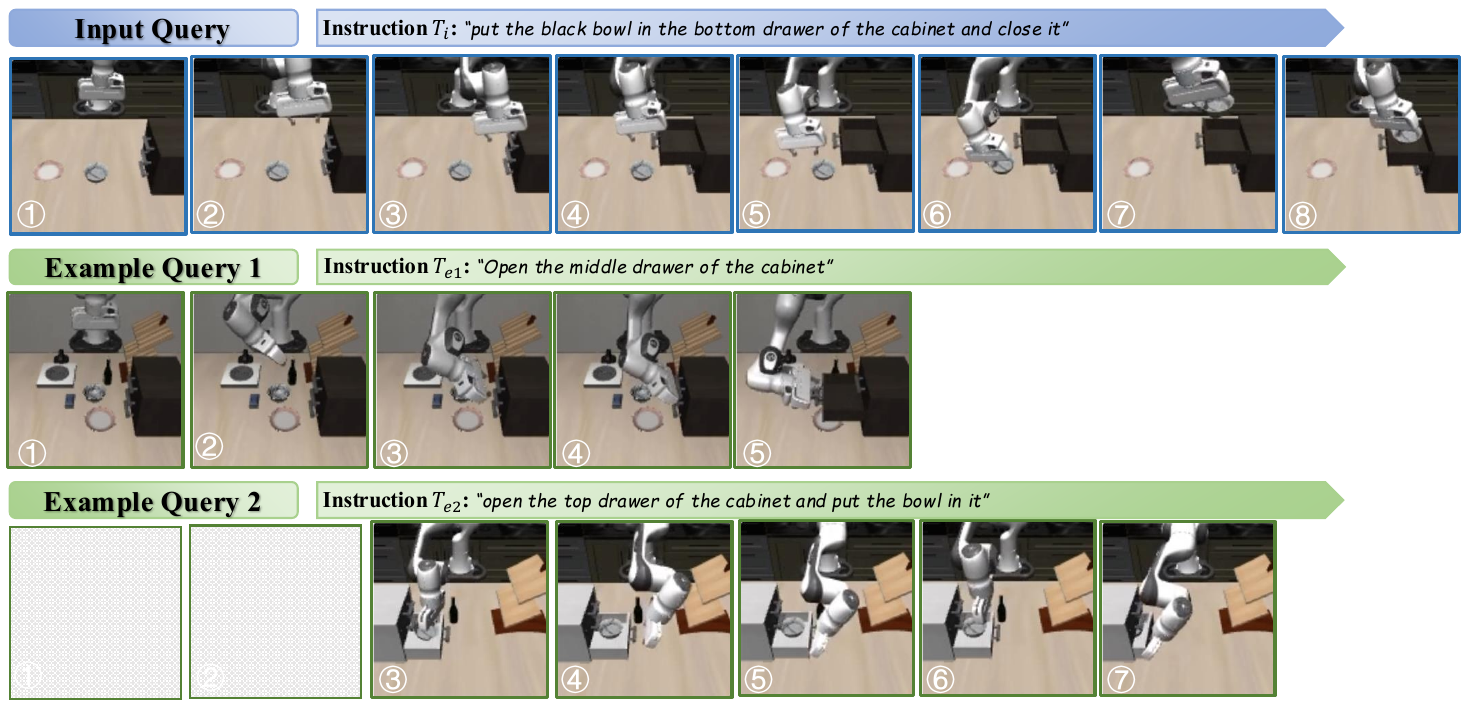}
\caption{\textbf{Qualitative visualization of phase-aligned retrieval.} The top row shows a long-horizon query rollout; the bottom rows show the references retrieved for different execution phases.}
\label{fig:qualitative}
\end{figure*}

\subsection{Baselines}


We compare ICI-VLA with VLA systems spanning discrete-token, generative-action, and specialized-control architectures: Octo~\cite{octo2024octo}, OpenVLA~\cite{kim2025openvla}, $\pi_0$-FAST~\cite{black2025pi0,pertsch2025fast}, MolmoAct~\cite{lee2025molmoact}, $\pi_0$~\cite{black2025pi0}, $\pi_{0.5}$-KI~\cite{physical2025pi05}, and OpenVLA-OFT~\cite{kim2025openvlaoft}. The reported baseline entries follow their original benchmark training and evaluation protocols and are included as reference points; accordingly, Table~\ref{tab:main_results} provides a benchmark-level rather than fully controlled comparison. 
VLA-0~\cite{goyal2025vla0} provides the closest policy-backbone reference, while its naive-ICL variant serves as the controlled comparison under our implementation.
REGENT~\cite{sridhar2025regent}, RICL~\cite{sridhar2025ricl}, and MEM~\cite{torne2026mem} are discussed conceptually because their evaluation protocols are not directly matched to ours.


\begin{table}[!t]
\centering
\resizebox{1.0\columnwidth}{!}{%
\begin{tabular}{l ccccc ccc}
\toprule
\multirow{2}{*}{\textbf{Configuration}} & \multicolumn{5}{c}{\textbf{LIBERO}} & \multicolumn{3}{c}{\textbf{RoboTwin 2.0}} \\
\cmidrule(lr){2-6} \cmidrule(lr){7-9}
 & Spat. & Obj. & Goal & Long & Avg. & Easy & Hard & Avg. \\
\midrule
Naive ICL (w/o Masking) & 72.8 & 76.5 & 73.1 & 63.5 & 71.5 & 18.2 & 2.0 & 10.7 \\
w/o DTW & 95.2 & 96.1 & 92.4 & 86.3 & 92.5 & 45.1 & 15.4 & 31.4 \\
w/o Semantic & 90.1 & 92.5 & 88.4 & 82.6 & 88.4 & 51.2 & 22.8 & 38.1 \\
\midrule
\rowcolor{gray!10} \textbf{Full (ours)} & \textbf{98.5} & \textbf{98.7} & \textbf{98.0} & \textbf{96.8} & \textbf{97.7} & \textbf{72.4} & \textbf{46.3} & \textbf{60.4} \\
\bottomrule
\end{tabular}%
}
\caption{\textbf{Component ablations.} Average success rate (\%). ``Naive ICL (w/o Masking)'' is the same configuration as the main-table baseline; DTW and Semantic denote the corresponding retrieval components.}
\label{tab:ablation}
\end{table}

\subsection{Main Results}

Table~\ref{tab:main_results} reports the simulation success rates. On LIBERO, ICI-VLA achieves an average success rate of 97.7\%, compared with the highest reported baseline average of 96.4\% from OpenVLA-OFT in this benchmark-level comparison. ICI-VLA records the highest listed values on Spatial, Goal, and Long, while remaining within 0.8 points of OpenVLA-OFT on Object. Its strongest relative result occurs on LIBERO-Long, reaching 96.8\% compared with the highest reported baseline value of 93.2\%. This result is consistent with the intended role of phase-aligned micro-demonstrations in supporting sustained multi-step coordination.

On RoboTwin 2.0, ICI-VLA achieves a 60.4\% average success rate, 19.3 points above the highest reported baseline value in Table~\ref{tab:main_results}. It also records the highest listed values on both Easy and Hard, with margins of 17.2 and 21.8 points, respectively. Because the external baseline entries follow their original evaluation protocols, these margins provide benchmark-level context rather than controlled paired evidence. Under our controlled implementation, however, VLA-0 with naive contextual demonstrations achieves only 10.7\%, whereas ICI-VLA reaches 60.4\% with the same policy backbone, trajectory partitions, optimization setting, and retrieved contexts. Together with the component ablations in Table~\ref{tab:ablation}, these results support the importance of both retrieval alignment and Target Action Masking for effectively exploiting contextual demonstrations.



\subsection{Ablation Studies}


To evaluate the contribution of each proposed module, we present component-level ablations in Table \ref{tab:ablation}.
Without Target Action Masking, average success decreases from 97.7\% to 71.5\% on LIBERO and from 60.4\% to 10.7\% on RoboTwin 2.0. Removing DTW ranking yields 92.5\% and 31.4\%, respectively, while removing semantic filtering yields 88.4\% and 38.1\%. Thus, every ablated variant underperforms the complete configuration under the reported protocol. However, these ablations quantify performance changes; they do not establish that masking induces explicit residual computation or that a particular internal shortcut causes every failure.
``VLA-0 w/ Naive ICL'' is trained and evaluated with the same policy backbone, trajectory partitions, optimization setting, and retrieved contexts as ICI-VLA, but with Target Action Masking disabled.

\begin{figure}[t!]
\centering
\includegraphics[width=\columnwidth]{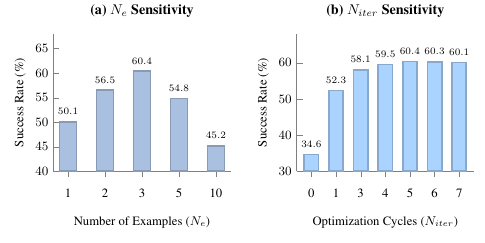}
\caption{\textbf{Sensitivity on RoboTwin 2.0.} (a) Number of retrieved demonstrations ($N_e$). (b) Number of retriever-mining cycles ($N_{iter}$).}
\label{fig:sensitivity}
\end{figure}




\subsection{Qualitative Analysis}

Figure~\ref{fig:qualitative} visualizes a representative execution of the compound instruction ``put the black bowl in the bottom drawer ... and close it.'' The RD-Encoder selects Reference~1 during drawer opening and switches to Reference~2 as the policy proceeds to bowl placement. This phase-dependent transition illustrates how the retriever supplies locally relevant context as a long-horizon task evolves; retrieval accuracy is evaluated quantitatively in the next section.





\begin{figure}[t]
\centering
\includegraphics[width=1.0\columnwidth]{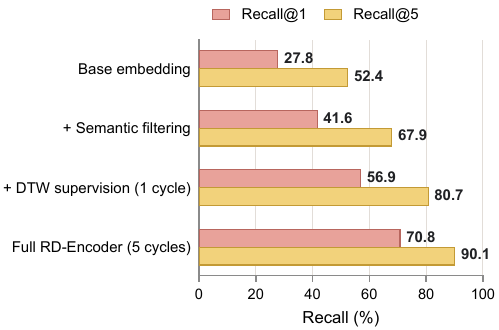}
\caption{\textbf{Direct retriever evaluation on RoboTwin 2.0.} Recall@$K$ is the percentage of queries whose offline DTW-defined positive appears among the top $K$ results.}
\label{fig:retriever_recall}
\end{figure}

\subsection{Direct Retriever Evaluation}

To isolate retrieval quality from action generation, we evaluate held-out RoboTwin 2.0 queries using the offline DTW-defined positive $P^{+}$ as the relevance target. The retriever ranks candidates from observable query inputs only and never receives the query's target action; Recall@$K$ measures whether $P^{+}$ appears among the top $K$ results. As shown in Figure~\ref{fig:retriever_recall}, Recall@1/Recall@5 increases from 27.8/52.4 for the base embedding to 41.6/67.9 after semantic filtering, 56.9/80.7 after one DTW-supervised mining cycle, and 70.8/90.1 for the full five-cycle RD-Encoder. Overall, the full retriever improves Recall@1 by 43.0 points and Recall@5 by 37.7 points, demonstrating progressively stronger retrieval as semantic filtering and DTW-mined supervision are introduced.

\subsection{Key Parameter Sensitivity}


Figure~\ref{fig:sensitivity} evaluates context size $N_e$ and retriever-mining cycles $N_{iter}$ on the RoboTwin 2.0 validation split. Performance peaks at $N_e=3$ (60.4\%) and decreases for both smaller and larger contexts, suggesting a trade-off between context coverage and irrelevant or conflicting information, although attention overload was not directly measured. Iterative mining raises performance from 34.6\% without optimization to 60.4\% after five cycles, with changes of at most 0.3 points through seven cycles. We therefore use three examples and five cycles---the smallest tested setting attaining the maximum---in the reported configuration, without tuning these parameters on the final test set.



\subsection{Real-World Evaluation}

\begin{figure}[t!]
\centering
\includegraphics[width=\columnwidth]{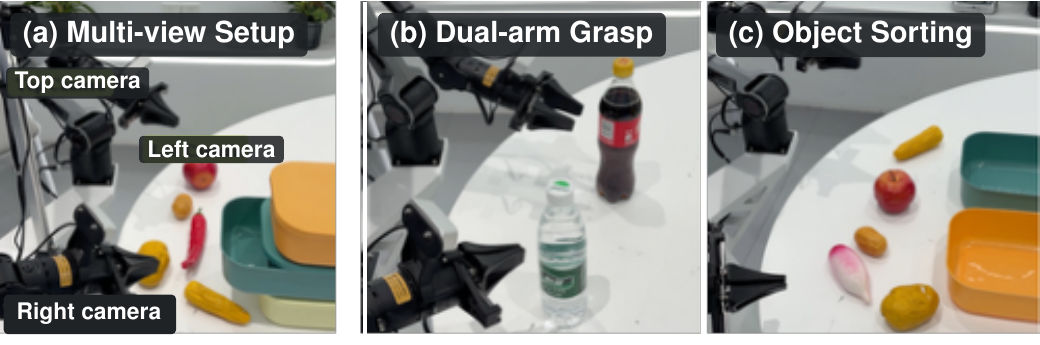}
\caption{\textbf{Physical dual-arm Aloha setup and representative tasks.} (a) Multi-view camera setup. (b) Dual-arm grasping. (c) Object sorting.}
\label{fig:real_world}
\end{figure}

We deploy ICI-VLA on a physical dual-arm Aloha system (Fig.~\ref{fig:real_world}) and evaluate Single-arm Grasp, Dual-arm Grasp, Drawer Placement, and Object Sorting under lighting variation, tabletop distractors, sensor noise, and contact dynamics. Evaluation objects, layouts, instructions, and trajectories are disjoint from policy training and the retrieval library, which contains approximately 1,000 other physical demonstrations. As reported in Table~\ref{tab:real_world}, over 250 rollouts per task (1,000 trials), ICI-VLA achieves 83.2\% success (95\% CI: 80.8--85.4), compared with 66.4\% (63.4--69.3) for $\pi_0$ and 63.0\% (60.0--65.9) for VLA-0.

\begin{table}[!t]
\centering
\resizebox{\columnwidth}{!}{%
\begin{tabular}{l cccc c}
\toprule
\textbf{Model} & \textbf{Single} & \textbf{Dual} & \textbf{Drawer} & \textbf{Sorting} & \makecell{\textbf{Avg.}\\\scriptsize 95\% CI} \\
\midrule
$\pi_0$ \cite{black2025pi0} & 78.4 & 56.8 & 62.0 & 68.4 & \makecell{66.4\\\scriptsize [63.4, 69.3]} \\
VLA-0 \cite{goyal2025vla0} & 75.2 & 52.4 & 58.8 & 65.6 & \makecell{63.0\\\scriptsize [60.0, 65.9]} \\
\midrule
\rowcolor{gray!10} \textbf{ICI-VLA (ours)} & \textbf{89.6} & \textbf{76.8} & \textbf{81.2} & \textbf{85.2} & \makecell{\textbf{83.2}\\\scriptsize \textbf{[80.8, 85.4]}} \\
\bottomrule
\end{tabular}
}
\caption{\textbf{Physical-task SR (\%).} Each task is evaluated over 250 rollouts; the final column gives the four-task average and its 95\% Wilson interval over 1,000 trials.}
\label{tab:real_world}
\end{table}

\section{Limitation and Conclusion}
\label{sec:conclusion}

We presented ICI-VLA, a retrieval-conditioned framework that combines spatiotemporal alignment and Target Action Masking to turn micro-demonstrations into effective context for robot control. With all parameters fixed at inference, ICI-VLA achieves 97.7\% on LIBERO, 60.4\% on RoboTwin 2.0, and 83.2\% across four physical tasks. These results demonstrate the potential of text-centric policies to adapt through well-aligned contextual experience across both simulation and physical settings.

ICI-VLA remains dependent on demonstration and planner coverage and incurs offline training and retrieval costs. Future work will improve retrieval efficiency, expand demonstration coverage, and investigate scalable cross-embodiment adaptation for more diverse real-world environments.

\bibliography{aaai2026}

\end{document}